\documentclass[letterpaper, 10 pt, conference]{ieeeconf}  

\IEEEoverridecommandlockouts                              
\usepackage{multirow}   
\pdfoutput=1
\usepackage{color}
\usepackage{graphicx}
\usepackage{caption}
\usepackage{amsmath}
\usepackage{amssymb}
\usepackage{xspace}
\usepackage{subfig}
\usepackage{booktabs}
\DeclareMathOperator*{\argmax}{arg\,max}
\DeclareMathOperator*{\argmin}{arg\,min}
\usepackage[export]{adjustbox}
\newcommand{\specialcell}[2][c]{%
  \begin{tabular}[#1]{@{}c@{}}#2\end{tabular}}

\newcommand{\set}[1]{ {\mathcal #1}\xspace}

\def\eg{\textit{e.g.}\xspace}
\def\ie{\textit{i.e.}\xspace}

\def\etal{\textit{et al.}\xspace}

\title{\LARGE \bf
Learning Deep NBNN Representations for Robust Place Categorization
}

\author{Massimiliano Mancini$^{1,2}$, Samuel Rota Bul\`o$^{2,3}$, Elisa Ricci$^{2,4}$, Barbara Caputo$^{1}$
\thanks{This work was partially supported by the ERC grant 637076 - RoboExNovo (B.C. ), and the CHIST-ERA project ALOOF (B.C.).}
\thanks{$^{1}$M. Mancini and B. Caputo are with University of Rome La Sapienza, Rome, Italy.
        {\tt\small \{mancini,caputo\}@dis.uniroma1.it}}%
\thanks{$^{2}$M. Mancini, S. Rota Bul\`o and E. Ricci are with Fondazione Bruno Kessler, Trento, Italy. {\tt\small \{rotabulo,eliricci\}@fbk.eu}}%
\thanks{$^{3}$S. Rota Bul\`o' is with Mapillary, Graz, Austria.}
\thanks{$^{4}$E. Ricci is with University of Perugia, Perugia, Italy.}%
}

\begin{document}
\begin{titlepage}
\null
\vfill
\renewcommand{\fboxsep}{10pt}
\fbox{\Large\begin{minipage}{\columnwidth}
\textbf{Disclaimer:}

This work has been accepted for publication in the IEEE Robotics and Automation Letters. 
\newline
\newline
\textbf{Copyright:} 
\newline
\copyright~2017 IEEE. Personal use of this material is permitted. Permission from IEEE must be obtained for all other uses,  in  any  current  or  future  media,  including  reprinting/  republishing  this  material  for  advertising  or promotional purposes, creating new collective works, for resale or redistribution to servers or lists, or reuse of any copyrighted component of this work in other works.
\newline
\end{minipage}}
\vfill
\clearpage
\end{titlepage}

\maketitle
\thispagestyle{empty}
\pagestyle{empty}

\begin{abstract}
This paper presents an approach for semantic place categorization using data obtained from RGB cameras. Previous studies on visual place recognition and classification have shown that, by considering features derived from pre-trained Convolutional Neural Networks (CNNs) in combination with part-based classification models, high recognition accuracy can be achieved, even in presence of occlusions and severe viewpoint changes. Inspired by these works,  we propose to exploit local deep representations, representing images as set of regions applying a Na\"{i}ve Bayes Nearest Neighbor (NBNN) model for image classification. As opposed to previous methods where CNNs are merely used as feature extractors, our approach seamlessly integrates the NBNN model into a fully-convolutional neural network.
Experimental results show that the proposed algorithm outperforms previous methods based on pre-trained CNN models and that, when employed in challenging robot place recognition tasks, it is robust to occlusions, environmental and sensor changes. 
\end{abstract}

\section{INTRODUCTION}

Recent years have seen the breakthrough
of mobile robotics into the consumer market. Domestic robots have become increasingly
common, as well as vehicles making use
of cameras, radar and other sensors to assist the driver.
An important aspect of human-robot interaction, is the ability of artificial agents to understand the way humans think and talk about abstract spatial concepts. For example,
a domestic robot may be asked to “clean the bathroom”, while a car may be asked to “stop at
the parking area”. Hence, a robot's definition of “bathroom”, or
“parking area” should point to the same set of places that a human would recognize as such.

The problem of assigning a semantic spatial label to an image has been extensively studied in the computer and robot vision 
literature \cite{oliva2001modeling,wu2011centrist,fazl2012histogram,pronobis2006discriminative,pronobis2009ijrr}. 
The most important challenges in identifying places 
come from the complexity of the concepts
to be recognized and from the variability of the conditions in which the images are captured. Scenes
from the same category may differ significantly, 
while images corresponding to different places
may look similar. 
The historical take on these issues has been to model the visual appearance of scenes considering a large variety of both global and local descriptors \cite{oliva2001modeling,wu2011centrist,fazl2012histogram,lazebnik2006beyond} and several (shallow) learning models (\textit{e.g.} SVMs, Random Forests).  
\begin{figure}[t]
  \centering
  \subfloat[Standard NBNN pipeline]{\includegraphics[width=0.22\textwidth]{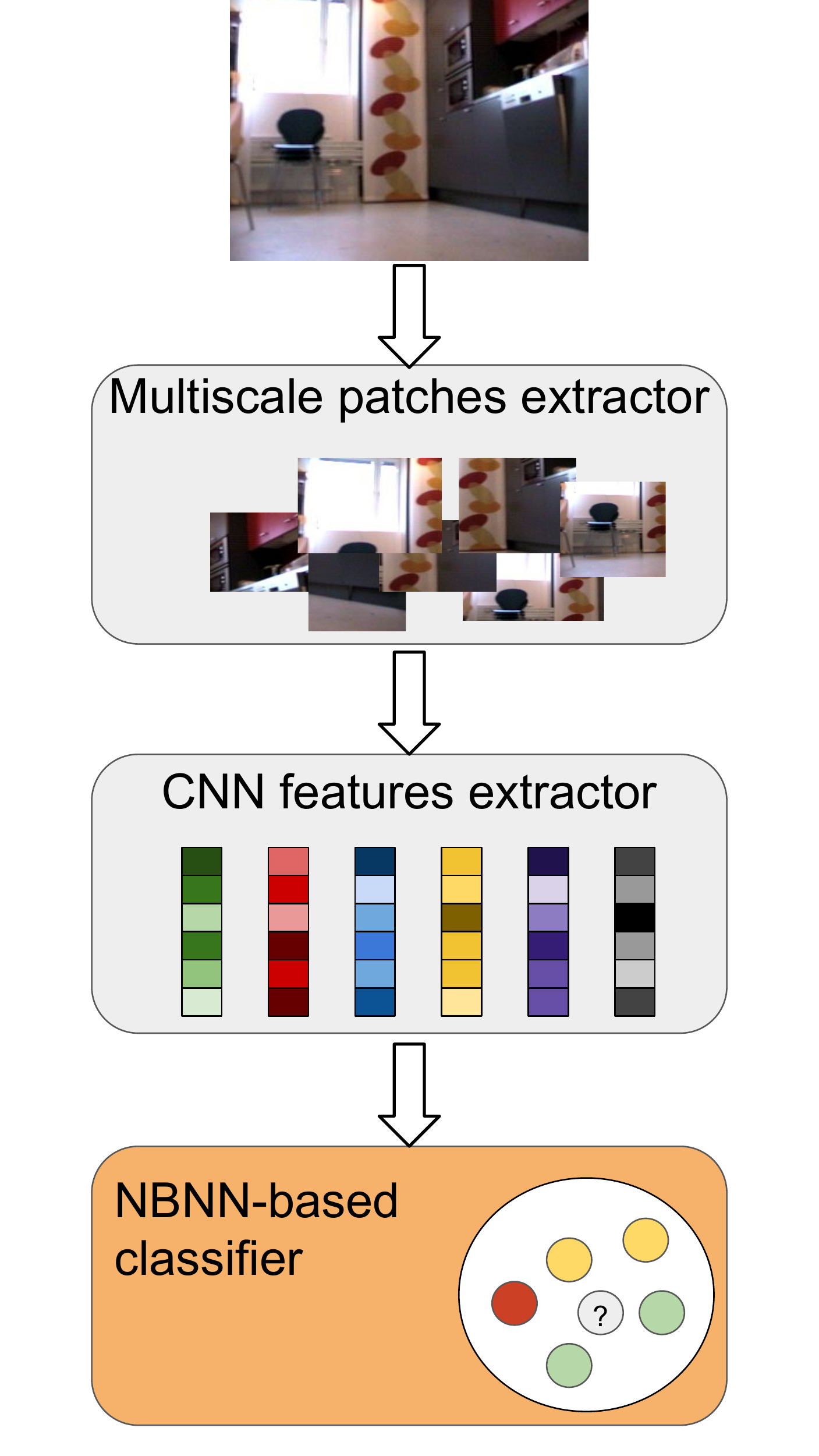}\label{fig:cnn_nbnn}}
  \hfill
  \subfloat[FullyConv-NBNN pipeline]{\includegraphics[width=0.22\textwidth]{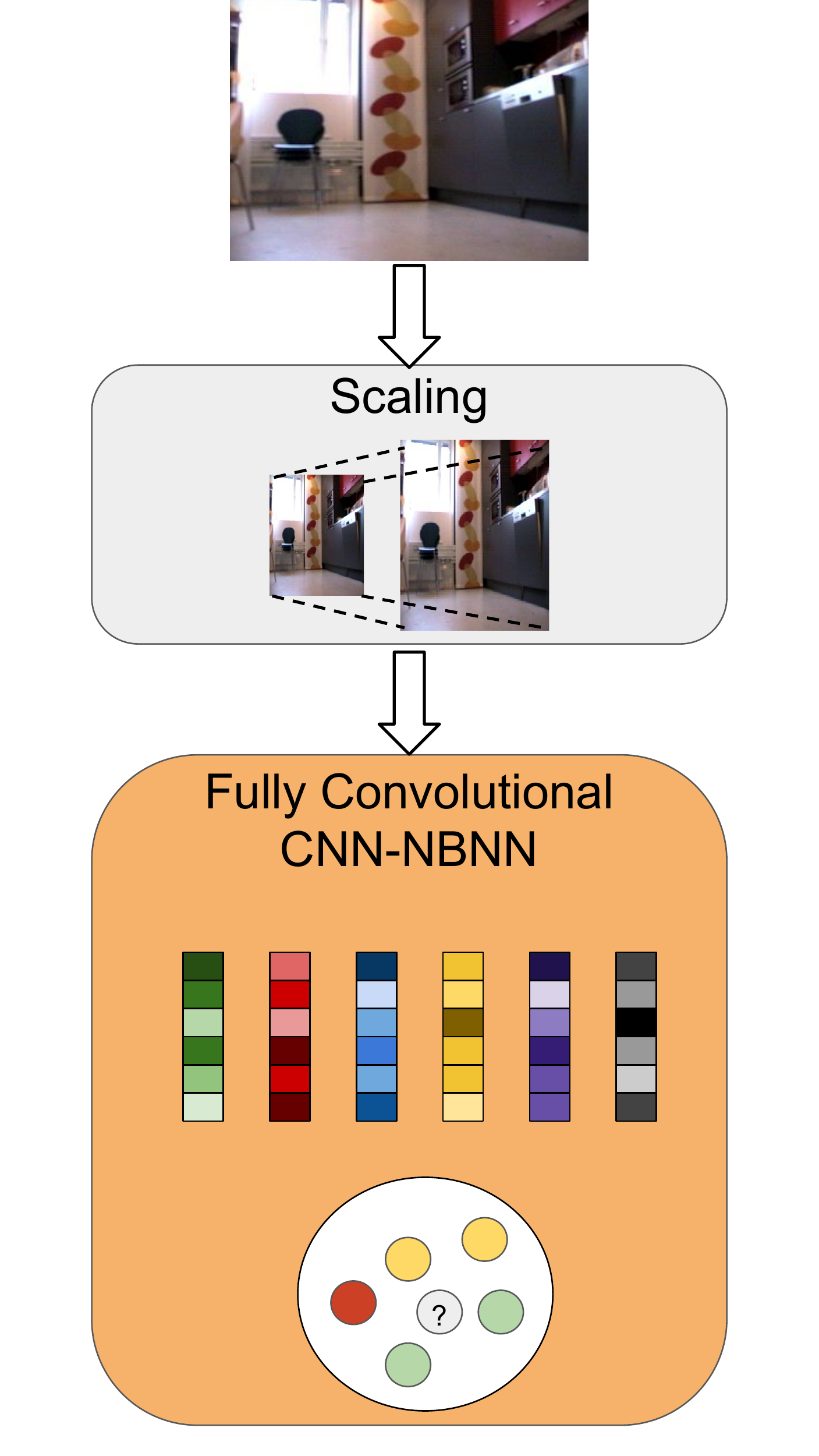}\label{fig:fullyconv}}
  \caption{The standard NBNN classification pipeline (a) versus the proposed model (b). The orange boxes indicate modules which involve a learning phase. Instead of extracting patches in a preprocessing step, we employ a fully-convolutional neural network, which automatically computes local features from the image. Moreover, features extraction and classifier modules are merged, allowing end-to-end training.}
  \label{fig:teaser}
  \vspace{-0.6cm}
\end{figure}

Since the (re-)emergence of Convolutional Neural Networks (CNNs), approaches based on learning deep representations have become  mainstream. 
Several works exploited deep models for visual-based scene classification and place recognition tasks, showing improved accuracy over traditional methods based on hand-crafted descriptors \cite{urvsivc2016part,sunderhauf2015performance,arroyo2016fusion,kanji2016self,neubert2015local}. Some of these studies \cite{urvsivc2016part,sunderhauf2015performance,neubert2015local} demonstrated the benefit of adopting a region-based approach (\textit{i.e.} considering only specific image parts) in combination with descriptors derived from CNNs, such as to obtain models which are robust to viewpoint changes and occlusions. With a similar motivation, lately several works in computer vision have attempted to bring back the notion of localities into deep networks, \textit{e.g.} by designing appropriate pooling strategies \cite{gong2014multi} or by casting the problem within the Image-2-Class (I2C) recognition framework \cite{kuzborskij2016naive}, with a high degree of success.
All these works decouple the choice of the significant localities from the learning of deep representations,
as the CNN feature extraction and the classifier learning are implemented as two separate modules.
This leads to two drawbacks: first, choosing heuristically the relevant localities means concretely cropping parts of the images before feeding them to the chosen features extractor.
This is clearly sub-optimal, and might turn out to be computationally expensive. Second, it would be desirable to fully exploit the power of deep networks by directly learning the best representations for the task at hand, rather than re-use architectures trained on general-purpose databases like ImageNet and passively processing patches from the input images without adapting its weights. Ideally, 
a fully-unified approach would guarantee more discriminative representations, resulting in higher recognition accuracy.

This paper contributes to this last research thread by addressing these two issues. We propose an approach for semantic place categorization which exploits local representations within a deep learning framework. Our method is inspired by the recent work~\cite{kuzborskij2016naive}, which demonstrates that, by dividing images into regions and representing them with CNN-based features, state-of-the-art scene recognition accuracy can be achieved by exploiting an I2C approach,  
namely a parametric extension of the Na\"{i}ve Bayes Nearest Neighbor (NBNN) model. 
Following this intuition, we propose a  deep architecture for semantic scene classification which seamlessly integrates the NBNN and CNN frameworks (Fig.~\ref{fig:teaser}). We automatize the multi-scale patch extraction process by adopting a fully-convolutional network \cite{long2015fully}, guaranteeing a significant advantage in terms of computational cost over two-steps methods. Furthermore, a differentiable counterpart of the traditional NBNN loss is considered to obtain an error that can be back-propagated to the underlying CNN layers, thus enabling end-to-end training.  
To the best of our knowledge, this is the first attempt to fully unify NBNN and CNN, building a deep version of Na\"{\i}ve Bayes Nearest Neighbor. 
We extensively evaluate our approach on several publicly-available benchmarks. Our results demonstrate the advantage of the proposed end-to-end learning scheme over previous works based on a two-step pipeline and the effectiveness of our deep network over state-of-the-art methods on challenging robot place categorization tasks.

\section{RELATED WORK}
\label{related}
In this section we review previous works on (i) visual-based place recognition and categorization and (ii) Na\"{i}ve Bayes Nearest Neighbor classification.

\subsection{Visual-based Place Recognition and Categorization}
In the last decade several works in the robotic community addressed the problem of developing robust place recognition \cite{cummins2008fab,kanji2015cross,lowry2016visual,sunderhauf2015performance,arroyo2016fusion} and semantic classification \cite{pronobis2006discriminative,costante2013transfer,urvsivc2016part} approaches using visual data.
In particular, focusing on place categorization from monocular images, earlier works adopted a two-step pipeline: first, hand-crafted features, such as GIST \cite{oliva2001modeling}, CENTRIST \cite{wu2011centrist}, CRFH \cite{pronobis2006discriminative} or HOUP \cite{fazl2012histogram}, are extracted from the query image, and then the image is classified into one of the predefined categories using a previously-trained discriminative model (\textit{e.g.}, Support Vector Machines). 
Similarly, earlier studies on visual-based place recognition and loop closing also considered hand-crafted feature representations \cite{cummins2008fab,kanji2015cross,ciarfuglia2012discriminative}.

More recently, motivated by the success of deep learning models in addressing visual recognition tasks \cite{krizhevsky2012imagenet}, robotic researchers have started to exploit feature representations derived from CNNs for both place recognition \cite{sunderhauf2015performance,arroyo2016fusion,neubert2015local} and semantic scene categorization \cite{urvsivc2016part} tasks. Sunderhau ̈\textit{et al.} \cite{sunderhauf2015performance} analyzed the performance of CNN-based descriptors with respect to viewpoint changes and time variations, presenting the first real-time place recognition system based on convolutional networks.
Arroyo \textit{et al.} \cite{arroyo2016fusion} addressed the problem of topological localization across different seasons and proposed an approach which fuses information derived from multiple convolutional layers of a deep architecture. Gout \textit{et al.} \cite{gout2017evaluation} evaluated the representational power of deep features for analyzing images collected by an autonomous surface vessel, studying the effectiveness of CNN descriptors in case of large seasonal and illumination changes. Ur{\v{s}}i{\v{c}} \textit{et al.} \cite{urvsivc2016part} proposed an approach for semantic room categorization: first, images are decomposed in regions and CNN-based descriptors are extracted for each region; then, a part-based classification model is derived for place categorization. Interestingly, they showed that their method outperforms traditional CNN architectures based on global representations \cite{krizhevsky2012imagenet}, as the part-based model guarantees robustness to occlusions and image scaling. Our work develops from a similar idea, but differently from \cite{urvsivc2016part} the deep network is not merely used as feature extractor and a novel CNN architecture, suitable to end-to-end training, is proposed. 

\subsection{Na\"{i}ve Bayes Nearest Neighbor Classification}
The NBNN approach has been widely adopted in the computer and robot vision community, as an effective method to overcome the limitations of local descriptor quantization and Image-2-Image recognition \cite{boiman2008defense}. Several previous studies have demonstrated that the I2C paradigm implemented by NBNN models is especially beneficial for generalization and domain adaptation \cite{tommasi2013frustratingly} and that, by adding a learning component to the non-parametric NBNN, performance can be further boosted \cite{fornoni2014scene}. 

Recent works have also shown that the NBNN can be successfully employed for place recognition and categorization tasks \cite{kuzborskij2016naive,kanji2015cross,kanji2016self}. Kanji \cite{kanji2015cross} introduced a NBNN scene descriptor for cross-seasonal place recognition. In a later work \cite{kanji2016self}, Kanji extended this approach by integrating CNN-based features and PCA, deriving a PCA-NBNN model for addressing the problem of self-localization in case of images with small view overlap. Kuzborskij \textit{et al.} \cite{kuzborskij2016naive} proposed 
a multi-scale parametric version of the NBNN classifier and demonstrated its effectiveness in combination with precomputed CNN descriptors for scene recognition. Our work is inspired by \cite{kuzborskij2016naive}. However, the proposed learning model is based on a fully-convolutional network which can be trained in an end-to-end manner. Therefore, it is significantly faster and more accurate than  \cite{kuzborskij2016naive}.

\section{Fully-Convolutional CNN-NBNL}
In this section we describe the proposed approach for semantic place categorization. 
As illustrated in Fig.~\ref{fig:teaser}, our method develops from the same idea of previous models based on local representations and CNN descriptors \cite{kuzborskij2016naive,urvsivc2016part}: images are decomposed into multiple regions (represented with CNN features) and a part-based classifier is used to infer the labels associated to places. However, differently from previous works, our approach unifies the feature extraction and the classifier learning phases, and we propose a novel CNN architecture which implements a part-based classification strategy. As demonstrated in our experiments (Sect.~\ref{experiments}), our deep network guarantees a significant boost in performance, both in term of accuracy and computational cost.
Since our framework is derived from previous works on NBNN-based methods \cite{boiman2008defense,fornoni2014scene,kuzborskij2016naive}, 
we first provide a brief description of these approaches (Sect.~\ref{sec:NBNN}-\ref{cnnnbnl}) and then we introduce the proposed fully-convolutional NBNN-based network (Sect.~\ref{sec:fcnnbnl}).

\begin{figure}[t]
\centering
\includegraphics[width=0.9\columnwidth]{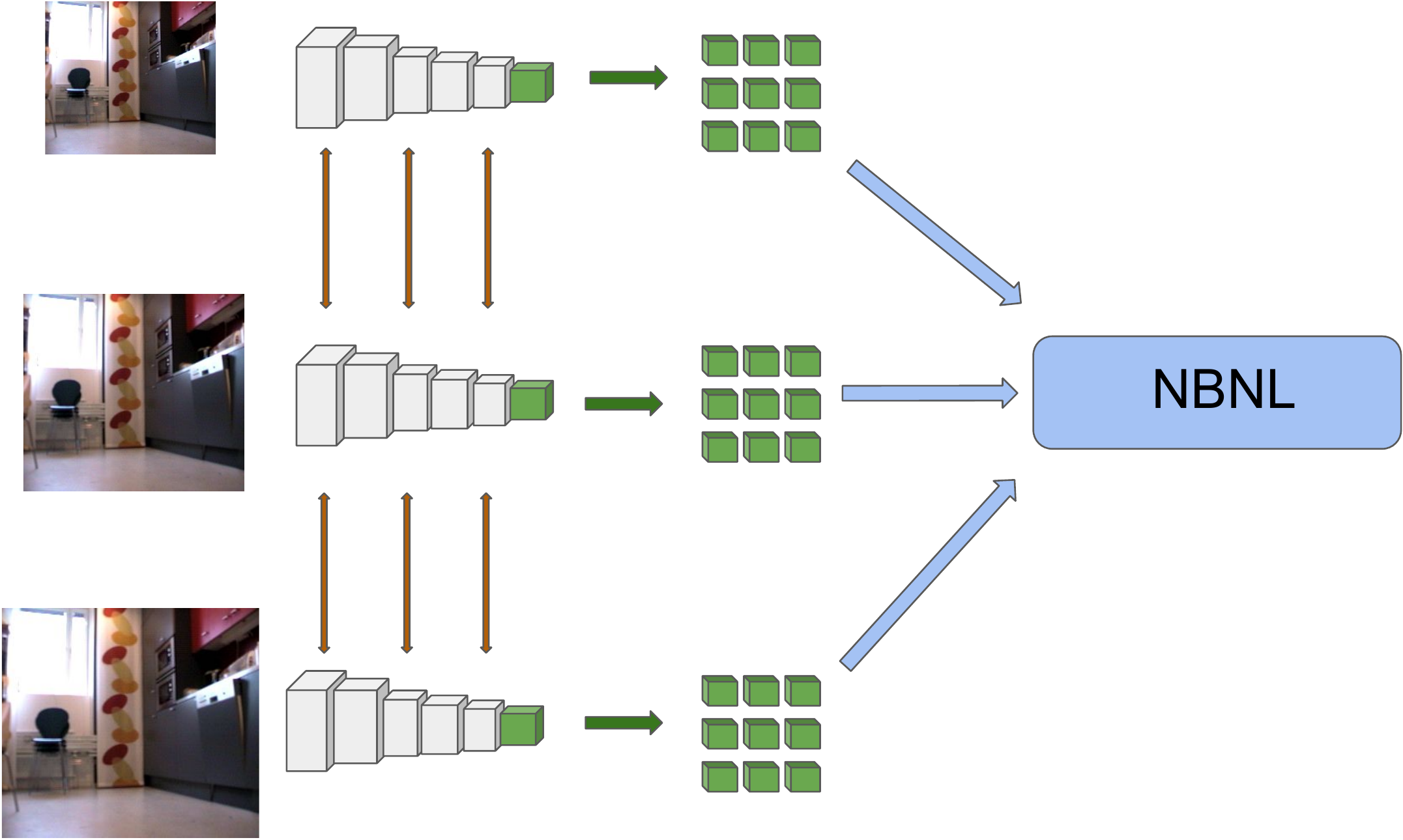}
    \caption{Simplified architecture of the proposed framework. The image is re-scaled to different sizes. The obtained images are fed in parallel to  multiple FC-CNNs with shared weights. From the networks, descriptors are extracted and used as input to the NBNL classifier.} 
   \label{fig:method}
   \vspace{-0.4cm}
\end{figure}

\subsection{Na\"{\i}ve Bayes Non-Linear Learning}
\label{sec:NBNN}
Let $\set X$ denote the set of possible images and let $\set Y$ be a finite set of class labels, indicating the different scene categories. The goal is to estimate a classifier $f:\set X\to\set Y$ from a training set $\set T\subset\set X\times\set Y$ sampled from the underlying, unknown data distribution.
The NBNN method~\cite{boiman2008defense} works under the assumption that there is a an intermediate Euclidean space $\set Z$ and a set-valued function $\phi$ that abstracts an input image $x\in\set X$ into a set of descriptors in $\set Z$, \ie $\phi(x)\subset\set Z$. For instance, the image could be broken into patches and a descriptor in $\set Z$ could be computed for each patch. Given a training set $\set T$, let $\Phi_y(\set T)$ be the set of descriptors computed from images in $\set T$ having labels $y\in\set Y$, \ie $\Phi_y(\set T)=\{\phi(x)\,:\,x\in\set X,(x,y)\in \set T \}$.
The NBNN classifier $f_\mathtt{NBNN}$ is given as follows:
\begin{equation}
f_\mathtt{NBNN}(x;\set T)=\argmin_{y\in\set Y}\sum_{z\in\phi(x)}d(z,\Phi_y(\set T))^2\,,
\label{eq:NBNN}
\end{equation}
where $d(x,\set S)=\inf\{\Vert z-s\Vert_2\,:\,s\in\set S\}$ denotes the smallest Euclidean distance between $z$ and an element of $\set S\subset\set Z$, or in other terms it is the distance between $z$ and its nearest neighbor in $\set S$.

Despite its effectiveness in terms of classification performance \cite{boiman2008defense}, $f_\mathtt{NBNN}$ has the drawback of being expensive at test time, due to the nearest-neighbor search. 
A possible way to reduce the complexity of this step consists in learning a small, finite set $\set W_y\subset\set Z$ of representative prototypes for each class $y\in\set Y$ to replace $\Phi_y(\set T)$. This idea was pursued by Fornoni \textit{et al.} \cite{fornoni2014scene} with a method named \textit{Na\"{\i}ve Bayes Non-Linear Learning} (NBNL). 
NBNL is developed from Eq.~\eqref{eq:NBNN} by replacing $\Phi_y(\set T)$ with the set of prototypes $\set W_y$ and by assuming $\set Z$ to be restricted to the unit ball. Under the latter assumption 
the bound $d(z,\set S)^2\geq 2-\omega(z,\set S)$ can be derived \cite{fornoni2014scene}, where:
\begin{equation}
\omega(z,\set S)= \left(\sum_{s\in\set S}|\langle z,s\rangle|_+^{q}\right)^{1/q}\,.
\label{eq:s}
\end{equation}
Here, $\langle\cdot\rangle$ denotes the dot product, $q\in[1,+\infty]$ and $|x|_+=\max(0,x)$. The NBNL classifier is finally obtained in the form given below by using the bound as a replacement of $d()^2$ in Eq.\eqref{eq:NBNN} (and after simple algebraic manipulations):
\begin{equation}
f_\mathtt{NBNL}(x;\set W)=\argmax_{y\in\set Y}\sum_{z\in\phi(x)}\omega(z,\set W_y)\,,
\label{eq:NBNL}
\end{equation}
where $\set W=\{\set W_y\}_{y\in \set Y}$ encompasses all the prototypes.

In order to learn the prototypes $\set W_y$ for each $y\in\set Y$, Fornoni \textit{et al.} did not consider $f_\mathtt{NBNL}$ as classifier and $\set T$ as training set, but they considered (only at training time) a classifier having the form $f(x)=\argmax_{y\in\set Y}\omega(z,\set W_y)$ and an extended training set $\{(z,y)\,:z\in\Phi_y(\set T), y\in\set Y\}$, where each descriptor extracted from an image is promoted to a training sample. In this way they derived the equivalent of a Multiclass Latent Locally Linear SVM (ML3) that is trained using the algorithm in~\cite{fornoni2013multiclass}. 

\subsection{CNN-NBNL}
\label{cnnnbnl}
Motivated by the robustness of NBNN/NBNL models and by the recent success of deep architectures in addressing challenging visual tasks, Kuzborskij \textit{et al.} \cite{kuzborskij2016naive} introduced an approach, named CNN-NBNL, which combines the NBNL and CNN frameworks. Their method is an implementation of NBNL, where $\phi(x)$ is obtained by dividing an image $x\in\set X$ into patches at different scales and by employing a pre-trained CNN-based feature extractor~\cite{jia2014caffe} to compute a descriptor for each patch. In formal terms, if $g_\mathtt{CNN}:\set X\to\set Z$ is the CNN-based feature extractor that takes an input image/patch and returns a single descriptor, then $\phi(x)$ (see, Sect.~\ref{sec:NBNN}) is given by 
\begin{equation}
\phi_\mathtt{CNN}(x)=\{g_\mathtt{CNN}(\hat x)\,:\,\hat x\in \text{patches}(x)\}\,,
\label{eq:phi-CNN}
\end{equation}
where $\text{patches}(x)\subset\set X$ returns a set of patches extracted from $x$ at multiple scales and reshaped to be compatible in terms of resolution with the input dimensionality required by the implementation of $g_\mathtt{CNN}$ (\eg CaffeNet~\cite{jia2014caffe} requires $227\times 227$).
To learn the prototypes $\set W_y$ in~\cite{kuzborskij2016naive} a training objective similar to~\cite{fornoni2014scene} is adopted, but the optimization is performed using a stochastic version of ML3 (STOML3) that better scales to larger datasets. At test time, $f_\mathtt{NBNL}$ defined as in Eq.~\ref{eq:NBNL} is used with $\phi$ replaced by $\phi_\mathtt{CNN}$.

 By moving from hand-crafted features to CNN-based features, the performance of the NBNL classifier improves considerably. Nonetheless, the approach proposed in~\cite{kuzborskij2016naive} has two limitations: 1) it requires the extraction of patches for each image as a pre-processing step, and CNN-features are extracted \emph{sequentially} from each patch; 2) the CNN architecture is used as a mere feature extractor and the method lacks the advantage of an end-to-end trainable system. The first limitation has a negative impact on the computation time of the method, while the latter makes way for further performance boosts.
 
\subsection{Fully-Convolutional CNN-NBNL}
\label{sec:fcnnbnl}
To overcome the two limitations of CNN-NBNL mentioned above, in this work we introduce a fully-convolutional version of CNN-NBNL that is end-to-end trainable (Fig.~\ref{fig:method}).

\vspace{3pt}\noindent\textbf{Fully-convolutional extension.}
Extracting patches at multiple scales and extracting CNN features independently for each of them is a very costly operation, which severely impacts training and test time. In order to perform a similar operation but with a limited impact on computation time, we propose to employ a Fully-Convolutional CNN (FC-CNN)~\cite{long2015fully} to simulate the extraction of descriptors from multiple patches over the entire image. A FC-CNN can be derived from a standard CNN by replacing fully-connected layers with convolutional layers. By doing so, the network is able to map an input image of arbitrary size into a set of spatially-arranged output values (descriptors). 
To cover multiple scales, we simply aggregate descriptors that are extracted with the FC-CNN from images at different resolutions. In this way, 
as the receptive fields of the FC-CNN remain the same, changing the scale of the input image induces an implicit change in the scale of the descriptors. 
The number of obtained descriptors per image depends on the image resolution and can in general be controlled by properly shaping the convolutional layers: for instance, by increasing the stride of the last convolutional layer it is possible to reduce the number of descriptors that the FC-CNN returns.

In the following, we denote by $g_\mathtt{FCN}(x;\theta)\subset\set Z$ the output of a FC-CNN parametrized by $\theta$ applied to an input image $x\in\set X$. As opposed to $g_\mathtt{CNN}$ defined in Sect.~\ref{cnnnbnl}, which returns a single descriptor, $g_\mathtt{FCN}(x;\theta)$ outputs a set of descriptors, one for each spatial location in the final convolutional layer of the FC-CNN. Each descriptor has a dimensionality that equals the number of output convolutional filters. We will also denote by $\eta(x)$ the number of descriptors that the FC-CNN generates for an input image $x$. Note that this number does not depend on the actual parametrization of the network, but only on its topology, which is assumed to be fixed, and on the resolution of the input image.

\begin{figure*}
\centering\includegraphics[width=\textwidth]{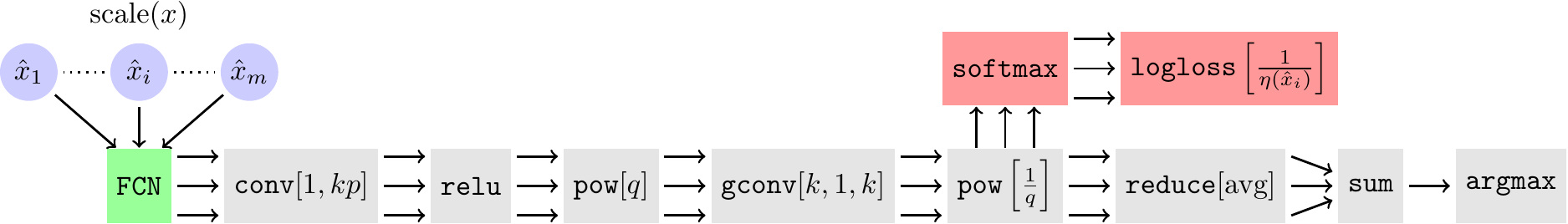}
\caption{Architecture of our fully-convolutional CNN-NBNL. 
We scale an input image $x\in\set X$ and obtain  $\{\hat x_1,\ldots,\hat x_m\}=\text{scale}(x)$. The scaled versions of $x$ are forwarded in parallel through the net.
The green block represents the FC-CNN. The gray blocks implement the NBNL classifier. The red blocks (top-left) are active only during training. Parameter $k$ represents the number of classes, $p$ the number of prototypes per class and $q$ the parameter in~Eq.\eqref{eq:s}. Further details about the building blocks are given hereafter.
\texttt{FCN} is a FC-CNN. \texttt{conv}[W,C] is a $W\times W$ convolutional layer with $C$ filters. \texttt{relu} applies the ReLu non-linearity to each element. \texttt{pow}[E] raises each element to the power of $E$. \texttt{gconv}[G,W,C] is a grouped $W\times W$ convolutional layer with $G$ groups and $C$ filters (the filters are filled and fixed with 1s; biases are omitted). \texttt{reduce}[avg] averages out the spatial dimensions. \texttt{sum} performs the element-wise sum of the incoming lines. \texttt{argmax} returns the index of the maximum element. \texttt{softmax} applies the softmax operator along the input channels, for each spatial entry of each input line. \texttt{logloss}[$\frac{1}{\eta(\hat x_i)}$] sums up the log-loss computed along the input channels of each spatial entry of each input line, and each input line $i$ is weighted by $\frac{1}{\eta(\hat x_i)}$. }
\label{fig:architecture}
\vspace{-15pt}
\end{figure*}

\vspace{3pt}\noindent\textbf{End-to-end architecture.} 
The NBNL classifier that we propose and detail below can be implemented using layers that are commonly found in deep learning frameworks and can thus be easily stacked on top of a FC-CNN (see, Fig.~\ref{fig:architecture}). By doing so, we obtain an architecture that can be trained end-to-end.

Given an input image $x\in\set X$, we create a set of $m$ scaled versions of $x$, which we denote by $\text{scale}(x)\subset\set X$.
Each scaled image $\hat x\in\text{scale}(x)$ is fed to the FC-CNN described before, yielding a set of descriptors $g_\mathtt{FCN}(\hat x;\theta)$. Instead of aggregating the descriptors from each scale, as done in Eq.~\eqref{eq:phi-CNN}, we keep them separated because they undergo a normalization step which avoids biasing the classifier towards scales that have a larger number of descriptors. 
The final form of our NBNL classifier is given by:
\begin{equation}
f_\mathtt{FCN\,NBNL}(x;\set W,\theta)=\argmax_{y\in\set Y}h(x;\set W_y,\theta)\,,
\end{equation}
where $h$ defined below measures the likelihood of $x$ given prototypes in $\set W_y$:
\begin{equation}
h(x;\set W_y,\theta)=\frac{1}{m}\sum_{\hat x\in\text{scale(x)}}\bar\omega(\hat x;\set W_y,\theta)
\end{equation}
and $\bar\omega$ is the scale-specific normalized score:
\begin{equation}
\bar\omega(\hat x;\set W_y,\theta)=\frac{1}{\eta(\hat x)}\sum_{z\in g_\mathtt{FCN}(\hat x;\theta)}\omega(z;\set W_y)\,.
\end{equation}
This normalization step is necessary to prevent scales that generate many descriptors to bias the final likelihood.

To train our network we define the following regularized empirical risk with respect to both the classifiers' parameters $\set W$ and the FC-CNN's parameters $\theta$:
\[
R(\set W,\theta;\set T)=\frac{1}{\set T}\sum_{(x,y)\in\set T}\ell(h(x;\set W,\theta), y) + \lambda \Omega(\set W,\theta)\,.
\]
Here, $h(x;\set W,\theta)=\{h(x;\set W_y,\theta)\}_{y\in\set Y}$, $\Omega$ is a  $\ell_2$-regularizer acting on all the networks' parameters, and $\ell(u,y)$ with $u=\{u_y\}_{y\in\set Y}$, $u_y\in\mathbb R$, is the following loss function:
\[
\ell(u,y)=-u_y+\log\sum_{y\in\set Y}e^{u_y}\,,
\]
obtained from the composition of the log-loss with the soft-max operator.

Following \cite{fornoni2014scene,kuzborskij2016naive} we actually do not minimize directly $R(\set W,\theta;\set T)$ as defined above, but replace the loss terms with the following upper-bound, which is obtained by application of Jensen's inequality:
\[
\ell(h(x;\set W,\theta), y)\leq 
\frac{1}{m}\sum_{\hat x\in\text{scale}(x)}\frac{1}{\eta(\hat x)}\sum_{z\in g_\mathtt{FCN}(\hat x;\theta)}\ell(\omega(z,\set W),y)\,,
\]
with $\omega(z,\set W)=\{\omega(z,\set W_y)\}_{y\in \set Y}$. This is equivalent to promoting descriptors to training samples, as in \cite{fornoni2014scene,kuzborskij2016naive}.

\section{EXPERIMENTAL RESULTS} 
\label{experiments}

In this section, we evaluate the performance of our approach. In Sect.~\ref{sceneexperiments} we compare against the method in \cite{kuzborskij2016naive}, demonstrating the advantages of our end-to-end learning framework. 
In Sect.~\ref{coldexp} we assess the effectiveness of the proposed approach for  the place categorization task, considering images acquired from different robotic platforms in various indoor environments, comparing with state-of-the-art 
approaches. Finally, we demonstrate the robustness of our model to different environmental conditions and sensors (Sect.~\ref{crossexp}) and to occlusions and image perturbations (Sect.~\ref{occexp}). Our evaluation has been performed using NVIDIA GeForce 1070 GTX GPU, implementing our approach with the popular Caffe framework \cite{jia2014caffe}.

\subsection{Comparison with Holistic and Part-based CNN models}
\label{sceneexperiments}
In a first series of experiments 
we demonstrate the advantages of the proposed part-based model and compare it with (i) its non end-to-end counterpart (\textit{i.e}, the CNN-NBNL method in \cite{kuzborskij2016naive}) and (ii) traditional CNN-based approaches not accounting for local representations. 
To implement \cite{kuzborskij2016naive}
following the original paper, we split the input image into multiple patches, extracting features from the last fully-connected layer of a pre-trained CNN. The patches were extracted at three different scales (32,64,128 pixels) after the original image was rescaled (longest side 200 pixels). We adopted the sparse protocol in \cite{kuzborskij2016naive}, based on which features from 100 random patches are extracted. The features are equally distributed between the three scales and an additional descriptor representing the full image is considered.
As representative for deep models based on holistic representations, we chose the 
successful approach of Zhou \textit{et al.} \cite{zhou2014learning,zhou2016places}: they pre-train a CNN on huge datasets (\textit{i.e.}, ImageNet \cite{deng2009imagenet}, Places \cite{zhou2014learning,zhou2016places} or both in the hybrid configuration) and used it as features extractor for learning a linear SVM model. Note that this is a strong baseline, widely used in the computer vision community for scene recognition tasks. 


To demonstrate the generality of our contribution, we tested all models considering three different base networks: 
the Caffe \cite{jia2014caffe} version of AlexNet \cite{krizhevsky2012imagenet}, VGG-16 \cite{simonyan2014very} and GoogLeNet \cite{szegedy2015going}. For AlexNet and VGG-16 we considered the networks pre-trained on both Places \cite{zhou2014learning,zhou2016places} and ImageNet \cite{deng2009imagenet} datasets (\textit{i.e}, the hybrid configuration). For GoogLeNet no pre-trained hybrid network was available, thus we took the model pre-trained on Places365. 
In order to fairly compare our model with the baseline method in \cite{kuzborskij2016naive}, our fully-convolutional network was designed to match the resolution of local patches adopted in \cite{kuzborskij2016naive}.
To accomplish this, since a 128x128 patch covers 64\% of a 200x200 image, we rescaled the input image such that the receptive fields 
correspond to approximately 64\% of the input (\ie 355 pxls for CaffeNet and 350 pxls for VGG and GoogLeNet). The other scale features were obtained by upsampling the image twice with a deconvolutional layer. We extracted 25 local features for the larger scale (128x128 pxls), 36 for the medium and 49 for the smallest, for a total of 110 local descriptors. These number of features were obtained by regulating the stride of the last layers of the network. 
As in \cite{kuzborskij2016naive}, we extracted features at the last fully-connected layer level, applying batch normalization \cite{ioffe2015batch} before the classifier. Since the datasets considered in our evaluation have small/medium dimensions, fine-tuning was performed only in the last two layers of the network.  
The networks were trained with a fixed learning rate which was decreased twice of a factor $0.1$. 
To decide the proper learning rate schedule and number of epochs, we performed parameters tuning on a separate validation set.  
As parameters of the NBNL classifier, we chose $k=10$ and $p=2$, applying a weight decay of $10^{-5}$ on the prototypes. 
Notice that in our model we considered 110 descriptors, while 100 
were used for the baseline method in~\cite{kuzborskij2016naive}. However, we experimentally verified that a difference of 10 descriptors does not influence performance. This confirms previous findings in~\cite{kuzborskij2016naive}, where Kuzborskij \etal also tested their approach with a dense configuration employing 400 
patches without significant improvements in accuracy over the sampling protocol. 

We performed experiments on three different datasets, previously used in~\cite{kuzborskij2016naive}: 
Sports8~\cite{li2007and}, Scene15 \cite{lazebnik2006beyond} and MIT67~\cite{quattoni2009recognizing}. The Sports8 dataset~\cite{li2007and} contains 8 different indoor and outdoor sport scenes (rowing, badminton, polo, bocce, snowboarding, croquet, sailing and rock climbing). The number of images per category ranges from 137 to 200. We followed the common experimental setting, taking 70 images per class for training and 60 for testing. The Scene15 dataset \cite{lazebnik2006beyond} is composed by different categories of outdoor and indoor scenes. It contains a maximum of 400 gray scale images per category. We considered the standard protocol, taking 100 images for training and 100 for testing for each class. 
The MIT67~\cite{quattoni2009recognizing} is a common benchmark for indoor scene recognition. It contains images of 67 indoor scenes, with at least 100 images per class. We adopted the common experimental setting, using 80 images per class for training and 20 for testing.
For each dataset we took 5 random splits reporting the results as mean and standard deviation. 

\begin{figure}[t]
  \centering
  \subfloat[\cite{kuzborskij2016naive}]{\includegraphics[width=0.24\textwidth]{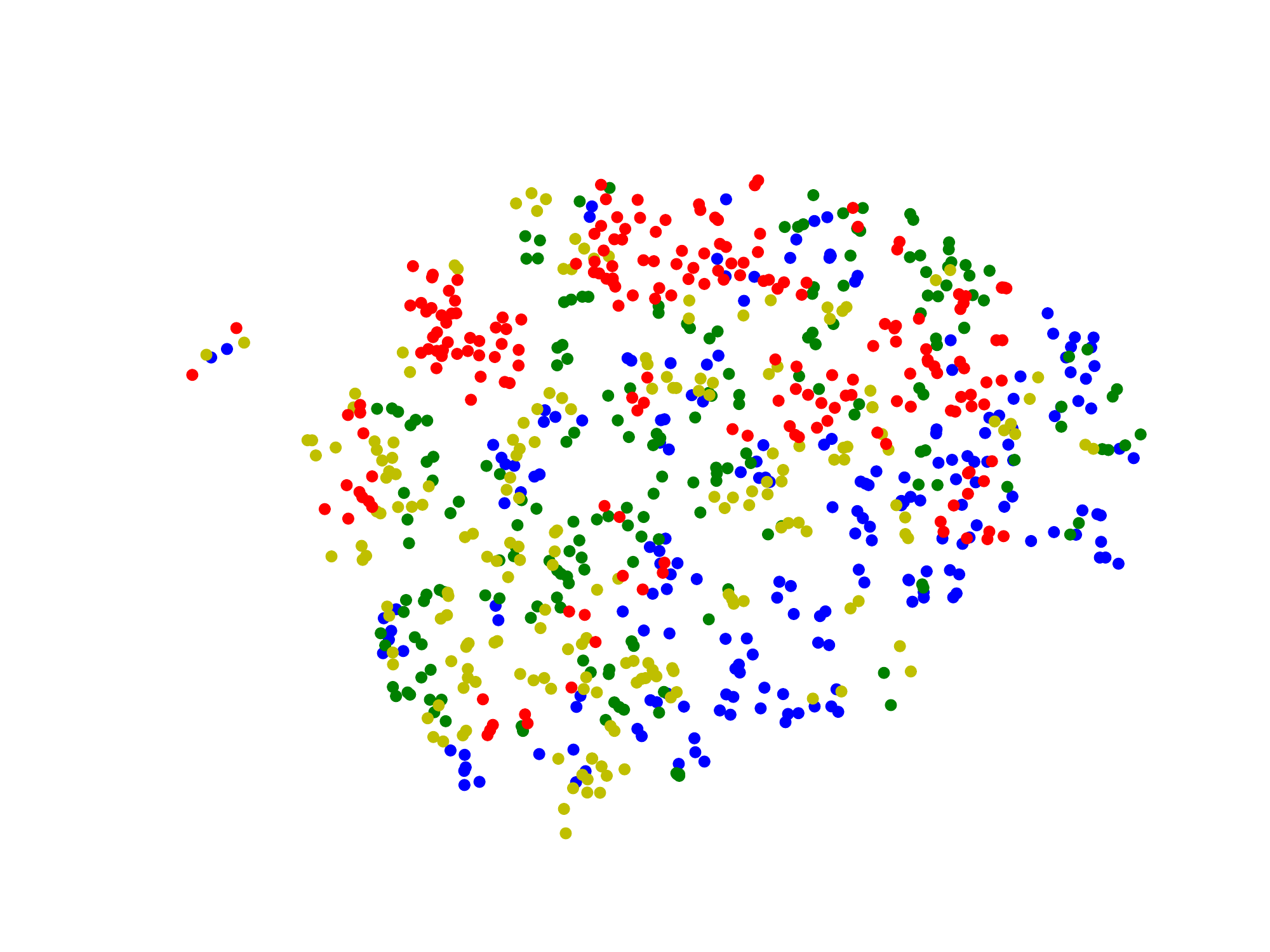}\label{fig:cnn_nbnl_tsne}} 
  \subfloat[Ours]{\includegraphics[width=0.24\textwidth]{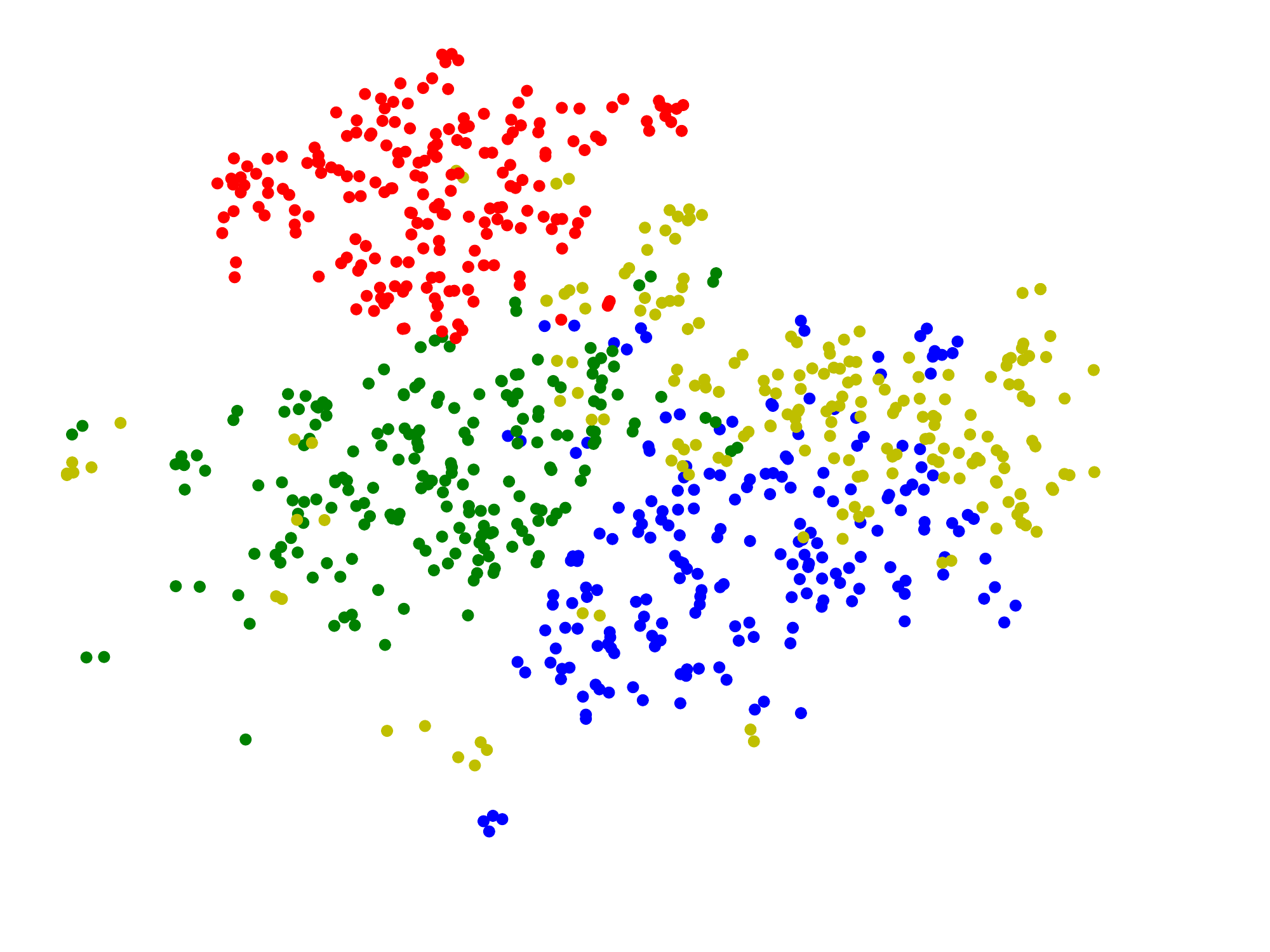}\label{fig:fullyconv_tsne}}
  \caption{t-SNE visualization of features extracted from 4 classes of the Scene15 dataset.  
  }
  \label{fig:tsne}
  \vspace{-0.5cm}
\end{figure}

Tab.~\ref{expScenes} shows the results of our evaluation. Mean and standard deviation are provided for our approach and \cite{kuzborskij2016naive}, while for the CNN models in~\cite{zhou2014learning,zhou2016places} we report results from the original papers. 
From the table it is clear that, for all base networks and datasets, our method outperforms the baselines. These results confirm the significant advantage of the proposed part-based approach over traditional CNN architectures which do not consider local representations. Moreover, our results show that our end-to-end training model guarantees an improvement in performance compared to its non end-to-end counterpart CNN-NBNL. This improvement is mostly due to the proposed end-to-end training strategy. A pre-trained network is able to extract powerful features, but they are not always discriminative when applied to specific tasks. On the other hand, end-to-end training allows to overcome this limitation by adapting the pre-trained features to a new target task, producing class discriminative representations. This is shown in Fig.~\ref{fig:tsne} where we plot the fc7 features extracted at scale 64x64 pixels (t-SNE visualizations~\cite{maaten2008visualizing}) with CNN-NBNL (Fig.~\ref{fig:tsne}.a) and with our approach (Fig.~\ref{fig:tsne}.b): while a pre-trained network fails at creating discriminative local features, our model is able to learn representations that cluster accordingly to class labels.   
\begin{table}[t]
			\caption{Comparing global and part-based CNN models. 
            \vspace{-5pt}} 
		\centering
		\scalebox{.9}{
		\begin{tabular}{| c | c | c | c | c |} 
			\hline
			Network& Method & Sports8 & Scene15 & MIT67\\ 
			\hline
			\multirow{3}{*}{\specialcell{AlexNet\\Hybrid}} &  \cite{zhou2014learning} & 94.22$\pm$0.78 &91.59$\pm$0.48& 70.8\\ 
			
			
			& \cite{kuzborskij2016naive}& $95.29\pm0.61$&$92.42\pm0.64$&$73\pm0.36$ \\ 
			
			 
			& Ours & \textbf{95.58 $\pm$ 0.58}&\textbf{93.63 $\pm$ 0.90}&\textbf{74.98 $\pm$ 0.78}  \\ 
			
			\hline
			\multirow{3}{*}{\specialcell{GoogLeNet\\Places365}}&   \cite{zhou2016places}& 91.00 &91.25& 73.30\\
			& \cite{kuzborskij2016naive}& $93.08\pm1.78$&$92.29\pm0.59$&$73.14\pm1.43$ \\ 			 
			& Ours & \textbf{94.46 $\pm$ 0.86}&\textbf{93.68 $\pm$ 0.57}&\textbf{80.55 $\pm$ 0.70}  \\
			
			\hline
			\multirow{3}{*}{\specialcell{VGG\\Hybrid}}
			&  \cite{zhou2016places} & 94.17 &92.12& 77.63\\ 			
			& \cite{kuzborskij2016naive}& $94.79\pm0.42$&$92.97\pm0.68$&$77.62\pm0.97$ \\			 
			& Ours & \textbf{97.04 $ \pm $ 0.27}&\textbf{95.12 $ \pm $ 0.41}&\textbf{82.49 $ \pm $ 1.35}  \\
			\hline
		\end{tabular}
        }
		\label{expScenes}
       \vspace{-10pt} 
\end{table}

\begin{figure}[t]
\centering
    \includegraphics[width=0.8\columnwidth]{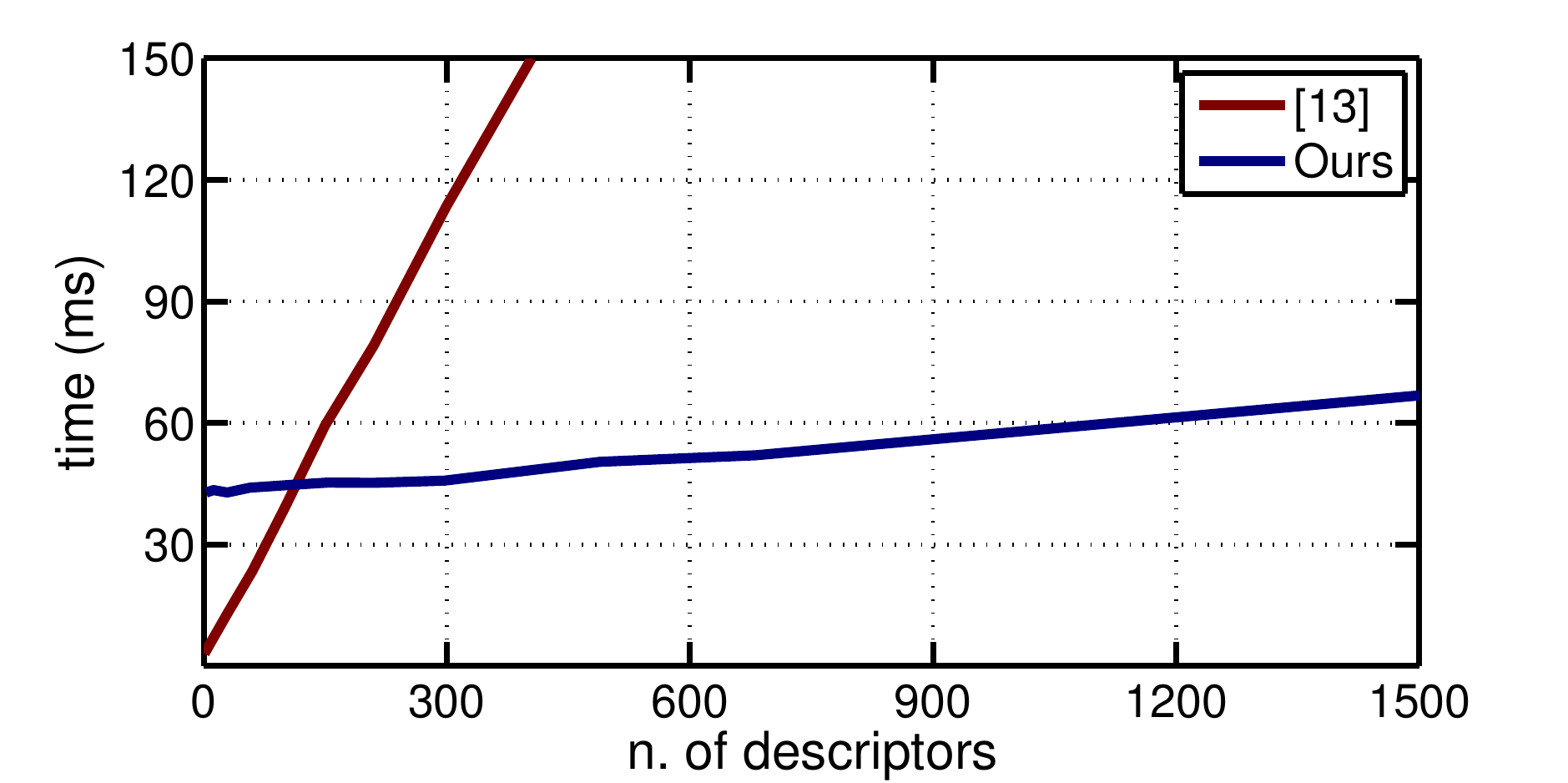}
    \caption{Computational time at varying number of descriptors. 
    } 
  \vspace{-0.4cm}
  \label{time_vs_patch}
\end{figure}

To further compare our approach and CNN-NBNL \cite{kuzborskij2016naive} we also analyzed the computational time required during the test phase to process an increasing number of patches. Fig.~\ref{time_vs_patch} report the results of our analysis: as expected, our fully-convolutional architecture is greatly advantageous over the CNN-NBNL model which extract local features independently patch-by-patch. 
We 
remark that reduced classification time is a fundamental for the adoption of the proposed model in robotic platforms operating in real environments. 


\subsection{Robot Place Categorization}
In this section we show the results of our evaluation when testing the proposed approach on publicly available robot vision datasets. These experiments aim at 
verifying the effectiveness of our fully-convolutional network and its robustness to varying environmental conditions and occlusions. 

\subsubsection{COLD dataset}
\label{coldexp}

We first tested our method on the COsy Localization Database (COLD) database \cite{pronobis2009ijrr}. This database contains three datasets of indoor scenes acquired in three different laboratories from different robots. The COLD-Freiburg contains 26 image sequences collected in the Autonomous Intelligent Systems Laboratory at the University of Freiburg with a camera mounted on an ActivMedia Pioneer-3 robot. COLD-Ljubljana contains 18 image sequences acquired from the camera of an iRobot ATRV-Mini platform at the Visual Cognitive Systems Laboratory of University of Ljubljana. In the COLD-Saarbr\"ucken an ActivMedia PeopleBot has been employed to gather 29 image sequences inside the Language Technology Laboratory at the German Research Center for Artificial Intelligence in Saarbr\"ucken. 

In our experiments we followed the protocol described in Rubio \textit{et al.} \cite{rubio2016comparison}, considering images of path 2 of each laboratory. These data depict significant changes with respect to illumination conditions and time of data acquisition. Using path 2, there are 9 categories for COLD-Freiburg, 6 for COLD-Ljubljana and 8 for COLD-Saarbr\"ucken. We trained and tested on data collected on the same laboratory, considering 5 random splits and reporting the average values. We compared our model with the methods proposed in \cite{rubio2016comparison}, since this work is one of the most recent studies adopting this dataset. In \cite{rubio2016comparison}, Rubio \textit{et al.} proposed to extract HOG features and to apply a dimensionality reduction technique 
before providing the features as input to different classifiers.  
As classifiers they considered linear SVM, 
Na\"{\i}ve Bayes (NB), Bayesian Network (BN) and the Tree Augmented Na\"{\i}ve Bayes (TAN). 
In our experiments, to train our model we adopted the same setting
described in Sect.~\ref{sceneexperiments}, fine-tuning the last two layers of the network.

The results are shown in Tab.~\ref{expCOLD}. Our model outperforms all the baselines in \cite{rubio2016comparison}, confirming the advantage of CNN-based approaches over traditional classifiers and hand crafted features. The high accuracy of our method also demonstrates that the proposed fully-convolutional network 
is highly effective at discerning among different rooms, even with significant lighting and environmental changes.

\begin{table}[t]
			\caption{Results on COLD dataset.\vspace{-5pt}} 
		\centering
		
		\begin{tabular}{| l | c | c | c |} 
			\hline
			Method & Freiburg & Saarbrücken & Ljubljana\\ 
			\hline HOG+SVM \cite{rubio2016comparison} & 46.5 &44.9& 66.2\\
			 HOG+NB \cite{rubio2016comparison} & 54.6 &52.9& 62.6\\
			 HOG+TAN \cite{rubio2016comparison} & 69.5 & 72.6& 75.2\\
			HOG+BN \cite{rubio2016comparison} & 82.3 &84.4& 88.5\\ 	 
			Ours& \textbf{95.2}&\textbf{97.3}&\textbf{99.2}  \\ 
			\hline
		\end{tabular}		   
		\label{expCOLD}
          \vspace{-0.4cm}
\end{table}
\subsubsection{KTH-IDOL dataset}
\label{crossexp}
To further assess the ability of the proposed method to 
generalize across different robotics platforms and illumination conditions, 
we performed experiments on the KTH Image Database for rObot Localization (KTH-IDOL) \cite{luo2006idol2}. This dataset contains 12 image sequences collected by two robots (Dumbo and Minnie) on 5 different rooms. The image sequences were collected along several days on three different illumination conditions: sunny, cloudy and night. Following \cite{wu2011centrist} we considered the first two sequences for each robot and weather condition, performing three different type of tests. First, we trained and tested using the same robot and same weather condition with one sequence used for training and the other for testing and vice-versa. As a second experiment, we used the same robot for training and testing, varying the weather conditions of the two sets. In the last experiment we trained the classifier with the same weather condition but testing it on a different robot. Notice that, differently from Sect.~\ref{coldexp}, in this case the illumination changes are not present in the training set. Our model is trained with the same setting of Sect.~\ref{sceneexperiments}. In this case, to reduce overfitting and improve the capability of our network, we apply data augmentation to the RGB channels, following the standard procedure introduced in \cite{krizhevsky2012imagenet}.

We compared our method with three state-of-the-art approaches: (i) \cite{pronobis2006discriminative} which used high dimensional histogram global features as input for a $\chi^2$ kernel SVM; (ii) \cite{wu2011centrist} which proposed the CENTRIST descriptor and performed nearest neighbor classification and (iii) \cite{fazl2012histogram} which used again the nearest neighbor classifier but with Histogram of Oriented Uniform Patterns (HOUP) as features. 

Tab.~\ref{expIDOL} shows the results of our evaluation. Our method outperforms all the baselines in the first and third series of experiments (same lighting). In particular, the large improvements in performance in the third experiment clearly demonstrates its ability to generalize over different input representations of the same scene, independently of the camera mounted on the robot. These results suggest that it should be possible to train offline our model and apply it on arbitrary robotic platforms. On the second experiment,
while the high classification accuracy demonstrates a significant robustness to lighting variations, our model achieves comparable performance with previous works, showing a small advantage of CNN representations over traditional methods in case of illumination changes. 


\begin{table}[t]
			\caption{Results on KTH-IDOL dataset (D and M denotes the names of the robot platforms Dumbo and Minnie).\vspace{-5pt}} 
		\centering		
		\begin{tabular}{| c | c | c || c | c | c | c | c |} 
			\hline
			Train & Test & Lighting &  \cite{pronobis2006discriminative}& \cite{wu2011centrist} &  \cite{fazl2012histogram} & Ours 
            \\ 
			\hline D & D & Same & 97.26 & 97.62 & 98.24 &\textbf{98.61} \\ 
			\hline M & M & Same & 95.51 & 95.35 & 96.61 & \textbf{97.32} \\ 
			\hline D & D & Diff & 80.55 & 94.98 & \textbf{95.76} &  94.17 \\ 
			\hline M & M & Diff & 71.90 & 90.17 & 92.01 & \textbf{93.62} \\ 
			\hline D & M & Same & 66.63 & 77.78 & 80.05 & \textbf{87.05} \\ 
			\hline M & D & Same & 62.20 & 72.44 & 75.43 & \textbf{88.51} \\ 
			\hline
		\end{tabular}		   
		\label{expIDOL}
          \vspace{-.9cm}
\end{table}

\subsubsection{Household room dataset}\label{occexp}

In the last series of experiments we tested the robustness of our model with respect to occlusions. We evaluate the performance of our approach on the recently introduced household room (or MIT8) dataset \cite{urvsivc2016part}. This dataset is a subset of MIT67 which contains 8 room categories: bathroom, bedroom, children room, closet, corridor, dining room, kitchen and living room. We used the setting provided in \cite{urvsivc2016part}, with 641 images for training and 155 for testing. The challenge proposed by Ur{\v{s}}i{\v{c}} \textit{et al.} \cite{urvsivc2016part} is to train the model on the original images and test its performances on various noisy conditions. The conditions are: occlusion in the center of the image, occlusion on the right border, occlusions from a person, addition of an outside border, upside down rotation and cuts on the top or right part of the image (inducing aspect ratio changes). All the test sets were produced following the protocol in \cite{urvsivc2016part}, apart from the person occlusions set provided directly by the authors.

We compare our approach with the part-based model developed by Ur{\v{s}}i{\v{c}} \textit{et al.} \cite{urvsivc2016part} and the global CNN-based model in \cite{zhou2014learning}. In \cite{urvsivc2016part} selective search is used to extract informative regions inside the image, which are then provided as input to a pre-trained CNN. From these features, exemplar parts are learned for each category and used by a part-based mixture model for the final classification. The standard hybrid CaffeNet \cite{zhou2014learning} is employed as CNN architecture. For a fair comparison we adopted the same base architecture, extracting features at the last fully-connected layer before the classifier. In this case we used images rescaled to 256x256 as input, upsampling them twice to obtain descriptors at multiple scales. We extracted 45 descriptors, 4 for the smallest scale (256x256), 16 for the medium and 25 for the largest. The training procedure is the one described in Sect.~\ref{sceneexperiments} and the same parameters are used for the NBNL classifier, with batch normalization applied to the last layer. We trained our model 10 times, computing the average accuracy. 

\begin{figure}[t]
\includegraphics[width=0.90\columnwidth,center]{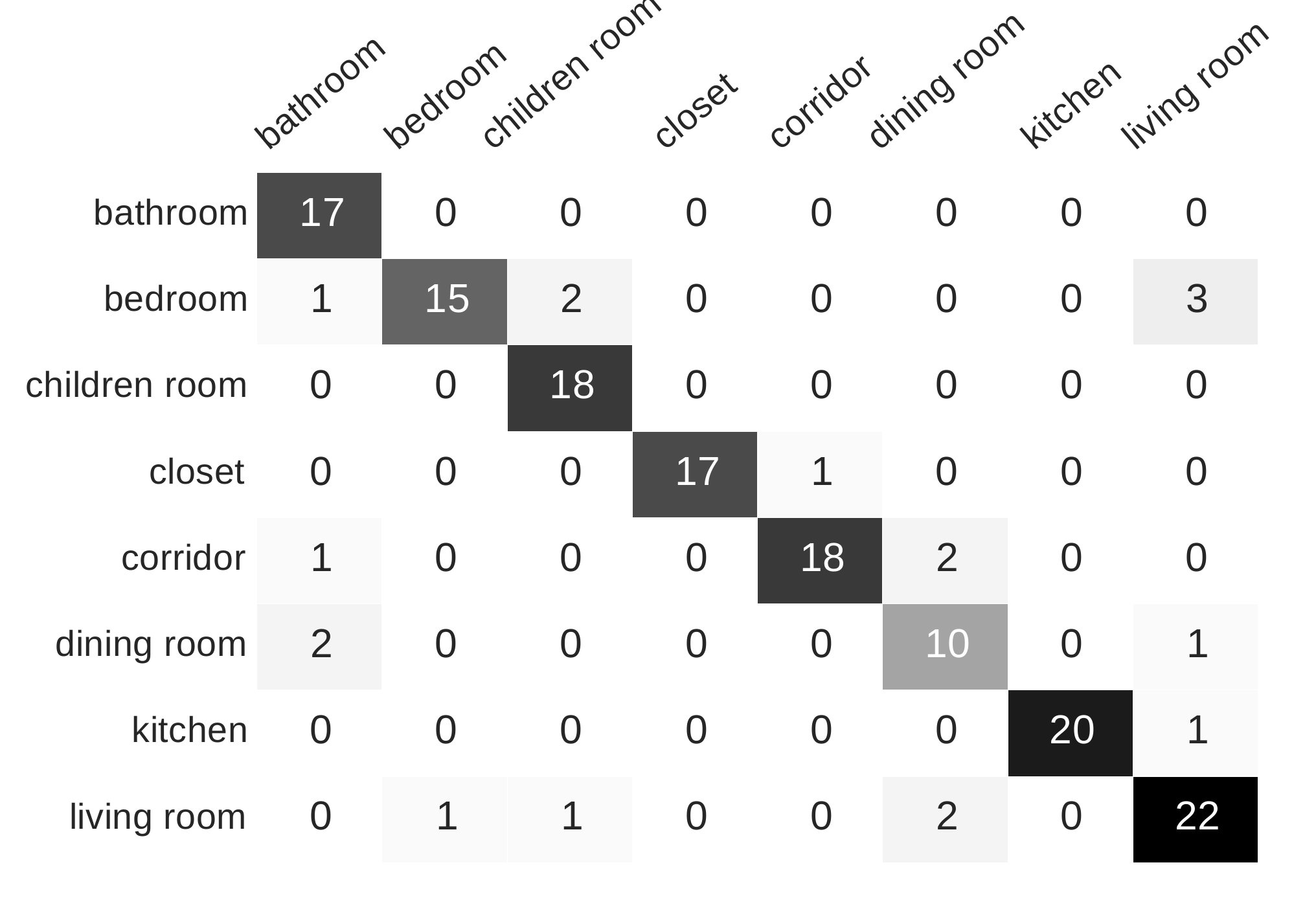}
    \caption{Confusion matrix obtained with our model classifying the original images of the MIT8 dataset.}   
  \label{fig:cm_mit8}
    \vspace{-0.4cm}
\end{figure}

The results of the evaluation are reported in Tab.~\ref{expMIT8}. As shown in the table, both our approach and the method in \cite{urvsivc2016part} achieve higher classification accuracy than the CNN model in \cite{zhou2014learning}, confirming the benefit of part-based modeling. It is interesting to compare our approach with \cite{urvsivc2016part}: while our framework guarantees better performances in certain conditions (\textit{e.g.} original frames, person occlusion), the method in \cite{urvsivc2016part} is more robust to changes of the aspect ratio (\textit{e.g.} cuts in the image) and scale (\textit{e.g.} outside border addition). Interestingly, when the occlusion is not created artificially obscuring patches (person occluder), our model achieves higher performance than \cite{urvsivc2016part}. Oppositely, in the case of the outside border experiments, almost half of the image is black and the real content reduces to a very small scale. In this (artificial) setting, \cite{urvsivc2016part} outperforms our model.
For sake of completeness, we also report the confusion matrix associated with our results on the original frames (Fig.~\ref{fig:cm_mit8}).


	\begin{table}[t]
			\caption{Results on MIT8 dataset.} 
		\centering
		
		\begin{tabular}{| l | c | c | c |} 
			\hline
			Experiment &  \cite{zhou2014learning} &  \cite{urvsivc2016part} & Ours \\ 
			\hline original  & 86.45 &85.16& \textbf{89.10} \\ 
			  outside border & 62.58 & \textbf{85.16} & 74.65  \\ 
			 black occluder, right & 78.71 & 80.00& \textbf{80.53} \\ 
			 black occluder, central & 61.94 &\textbf{69.68}& 67.74  \\ 
			 person occluder, central & 59.35&68.39& \textbf{72.45} \\ 
			 cut right half & 62.58 &64.52 &\textbf{65.16} \\ 
			 cut top half & 52.26 &\textbf{68.39}& 63.16 \\ 
			 upside down & 52.26 &59.35& \textbf{63.94} \\ 
			\hline
		\end{tabular}		   
		\label{expMIT8}
          \vspace{-0.3cm}
\end{table}

\section{CONCLUSIONS}
We presented a novel deep learning architecture for addressing the semantic place categorization task. By seamlessly integrating the CNN and NBNN frameworks, our approach permits to learn local deep representations, enabling robust scene recognition. 
The effectiveness of the proposed method is demonstrated on various benchmarks. We show that our approach outperforms traditional CNN baselines and previous part-based models which use CNNs purely as features extractors. In robotics scenarios, our deep network achieves state-of-the-art results on three different benchmarks, demonstrating its robustness to occlusions, environmental changes and different sensors. As future work, we plan to extend this model in order to handle multimodal inputs (\textit{e.g.} considering range sensors in addition to RGB cameras). 






\vspace{-10pt}
\bibliographystyle{IEEEtran}
\bibliography{IEEEabrv,iros}

\end{document}